\documentclass[a4paper]{article}

\usepackage{INTERSPEECH2016}

\usepackage{graphicx}
\usepackage{amssymb,amsmath,bm}
\usepackage{textcomp}

\ninept

\title{Active and Semi-Supervised Learning in ASR: \\ Benefits on the Acoustic and Language Models}

\makeatletter
\def\name#1{\gdef\@name{#1\\}}
\makeatother \name{{\em Thomas Drugman, Janne Pylkk\"{o}nen, Reinhard Kneser}}

\address{Amazon \\
  {\small \tt drugman@amazon.com, jannepyl@amazon.com, rkneser@amazon.com}}

\begin{document}

  \maketitle
  \begin{abstract}
The goal of this paper is to simulate the benefits of jointly applying active learning (AL) and semi-supervised training (SST) in a new speech recognition application. Our data selection approach relies on confidence filtering, and its impact on both the acoustic and language models (AM and LM) is studied. While AL is known to be beneficial to AM training, we show that it also carries out substantial improvements to the LM when combined with SST. Sophisticated confidence models, on the other hand, did not prove to yield any data selection gain. Our results indicate that, while SST is crucial at the beginning of the labeling process, its gains degrade rapidly as AL is set in place. The final simulation reports that AL allows a transcription cost reduction of about 70\% over random selection. Alternatively, for a fixed transcription budget, the proposed approach improves the word error rate by about 12.5\% relative.
  \end{abstract}

\section{Introduction}\label{sec:intro}
This paper aims at the problem of identifying the best approach for jointly selecting the data to be labelled, and maximally leveraging the data left as unsupervised. Our application targets voice search as used in various Amazon products. Because speech data transcription is a time-consuming and hence costly process, it is crucial to find an optimal strategy to select the data to be transcribed via active learning. In addition, the unselected data might also be helpful in improving the performance of the ASR system by semi-supervised training. As will be shown in this paper, such an approach allows to reduce the transcription cost dramatically while enhancing the customer's experience.

Active Learning (AL) refers to the task of minimizing the number of training samples to be labeled by a human so as to achieve a given system performance \cite{Cohn}. Unlabeled data is processed and the most informative examples with respect to a given cost function are then selected for human labeling. AL has been addressed for ASR purpose across various studies, which mainly differ by the measure of informativeness used for data selection. First attempts were based on so-called confidence scores \cite{Riccardi} and on a global entropy reduction maximization criterion \cite{Yu}. In \cite{Hamanaka}, a committee-based approach was described. In \cite{Huang}, a min-max framework for selecting utterances considering both informativeness and representativeness criteria was proposed. This method was used in \cite{Itoh} together with an N-best entropy based data selection. Finally, the study in \cite{Fraga} found that HMM-state entropy and letter density are good indicators of the utterance informativeness. Encouraging results were reported from the early attempts \cite{Riccardi, Yu} with a 60\% reduction of the transcription cost over Random Selection (RS).

In this paper, we focus on conventional confidence-based AL as suggested in \cite{Riccardi}, although other studies \cite{Yu,Itoh,Fraga} have shown some improvement over it. It is however worth highlighting that the details of the baseline confidence-based approach were not always clearly described, and that subsequent results were not in line with those reported in \cite{Riccardi}. First, various confidence measures can be used in ASR. A survey of possible confidence measures is given in \cite{Jiang} and several techniques for confidence score calibration have been developed in \cite{Yu_confidence}. Secondly, there are various possible ways of selecting data based on the confidence scores. 

Semi-Supervised Training (SST) has also recently received a particular attention in the ASR literature. A method combining multi-system combination with confidence score calibration was proposed in \cite{Huang_SST}. A large-scale approach based on confidence filtering together with transcript length and transcript flattening heuristics was used in \cite{Kapralova}. A cascaded classification scheme based on a set of binary classifiers was proposed in \cite{Li}. A reformulation of the Maximum Mutual Information (MMI) criterion used for sequence-discriminative training of Deep Neural Networks (DNN) was described in \cite{Manohar}. A shared hidden layer multi-softmax DNN structure specifically designed for SST purpose was proposed in \cite{Su}. The way unsupervised data\footnote{In this paper, "unsupervised data" refers to transcriptions automatically produced by the baseline ASR. In literature, training with automatic transcriptions produced by a supervised ASR system is sometimes referred to as "semi-supervised training", but we reserve the latter term to situations where both manual and automatic transcriptions are used together.} is selected in this paper is inspired from \cite{Kapralova}, as it is based on confidence filtering with possible additional constraints on the length and frequency of the transcripts. 

This paper aims at addressing the following questions whose answer is left either open or unclear with regard to the literature: \emph{i)} Do more sophisticated confidence models help improving data selection? \emph{ii)} Is AL also beneficial for LM training, and if so to which extent? \emph{iii)} How do the gains of AL and SST scale up when more and more supervised data is transcribed? \emph{iv)} Are the improvements similar after cross-entropy and sequence-discriminative training of the DNN AM?

In most of existing AL and SST studies (e.g. \cite{Riccardi, Yu, Itoh, Fraga, Manohar, Su}), the Word Error Rate (WER) typically ranges between 25 and 75\%. The baseline model in the present work has a WER of about 12.5\%, which makes the application of AL and SST on an industrial task even more challenging. 

The paper is structured as follows. Section \ref{sec:method} presents the approach studied throughout our experiments. Experimental results are described in Section \ref{sec:experiments}. Finally, Section \ref{sec:conclu} concludes the paper.

\section{Method}\label{sec:method}

Our method relies heavily on confidence-based data selection. Because confidence scores play an essential role, several confidence models have been investigated. They are described in Section \ref{ssec:conf}. The technique of data selection is presented in Section \ref{ssec:selection}. Details about AM and LM training are then provided respectively in Sections \ref{ssec:AM} and \ref{ssec:LM}.

\subsection{Confidence modeling}\label{ssec:conf}

As mentioned in the introduction, there are various confidence measures available \cite{Jiang, Yu_confidence, Senone_confidence}. First of all, confidence measures can be estimated at the token and utterance levels. The conventional confidence score at the token level is the token posterior from the confusion network \cite{Jiang}. It is however in practice a poor estimate of the actual probability of the token being correct, and is therefore lacking interpretability. This was addressed in \cite{Yu_confidence} by calibrating the scores using a maximum entropy model, an artificial neural network, or a deep belief network.

In this paper, confidence score normalization is performed to match the confidences with the observed probabilities of words being correct, using one of the two following methods: a piecewise polynomial which maps the token posteriors to confidences, or a linear regression model with various features such as the token posteriors, the word accuracy priors, the number of choices in the confusion network, and the number of token nodes and arcs explored. These two models are trained on an in-domain held-out data set.

As our data selection method processes utterances, it is necessary to combine the scores from the various tokens to get a single confidence measure at the utterance level. Conventional approaches encompass using an arithmetic or geometrical mean rule. In addition, we have also considered training a Multi-Layer Perceptron (MLP) to predict either the WER or the Sentence Error Rate (SER). The MLP took as input a vector consisting of statistics of the tokens: number of tokens, min/max and mean values of their posteriors.

\subsection{Data Selection}\label{ssec:selection}
Because the DNN we use for AM is a discriminative model, the selection of supervised data for AM purpose consists in maximizing the informativeness of the chosen utterances. Intuitively, this translates to selecting utterances with low confidence scores. Different settings of the confidence filter will be investigated in our experiments. Besides, we consider also filtering out too short utterances.

The selection of unsupervised data requires to find a balance between the informativeness and the quality of the automatic transcripts. This latter aspect imposes to retain only high confidence scores, as errors in the transcripts can be harmful to the training (particularly if it is sequence-discriminative \cite{Manohar}). As suggested in \cite{Kapralova}, utterance length and frequency filtering are additionally applied to flatten the data.

\subsection{AM training}\label{ssec:AM}

Our AM is a conventional DNN \cite{Hinton} made of 4 hidden layers containing 1536 units each. A context-dependent GMM is first trained using the Baum-Welch algorithm and PLP features. The size of the triphone clustering tree is about 3k leaves. The GMM is used to produce the initial alignment of the training data and define the DNN output senones. Our target language in this study is German, but we decided to apply transfer learning \cite{TransferLearning} by initializing the hidden layer weights from a previously-trained English DNN. The output layer was initialized with random weights. The input features are 32 standard Mel-log filter bank energies, spliced by considering a context of 8 frames on each side, therefore resulting in 544 dimensional input features.

The training consists of 18 epochs of frame-level cross-entropy (XE) training followed by boosted Maximum Mutual Information (bMMI) sequence-discriminative training \cite{bMMI}. The Newbob algorithm is used as Learning Rate (LR) scheduler during XE training. The learning rate for bMMI was optimized using a held-out development set. The resulting DNN is used to re-align the data and the same procedure of DNN training starting from transfer learning is applied again. The baseline model on the 50 initial hours was obtained in this way. For the next models which ingest additional supervised and/or unsupervised data, the baseline model is used to get the alignments, and the training procedure starting from transfer learning is performed.

\subsection{LM training}\label{ssec:LM}

Our LM is a linearly interpolated trigram model consisting of 9 components. The most important one (with interpolation weight $>0.6$) is trained on the selected supervised and unsupervised data. For the remaining components, we consider a variety of Amazon catalogue and text search data relevant for the voice search task. All component models are 3-gram models trained with modified Kneser-Ney smoothing \cite{Kneser}. The interpolation parameters are optimized on a held-out development set. The size of the LM is finally reduced using entropy pruning \cite{Stolcke}.

\section{Experiments}\label{sec:experiments}

The aim of our experiments is to simulate the possible gains obtained by AL and SST for a new application. For this simulation, we had about 600 hours of transcribed voice search data in German at our disposal. From this pool, 50 hours are first randomly selected to build the baseline AM and LM. These models are then used to decode the remaining 550 hours. The confidence models described in Section \ref{ssec:conf} and previously trained on a held-out set are employed so that each utterance in the 550h selection pool is assigned one confidence score (per confidence model). From the selection pool, the supervised data is selected first via conventional RS or via AL. Utterances which were left over are considered as unsupervised data for SST. The evaluation is carried out on a held-out dataset of about 8 hours of in-domain data. A speaker overlap with the training set is possible but the large number of speakers diminishes its potential effect. Our target metric is the standard WER.



In the next sections the results of the experiments are presented. Section \ref{ssec:conf_results} investigates the influence of the confidence model on data selection. The impact of AL and SST on both the AM and the LM is studied in Sections \ref{ssec:AM_results} and \ref{ssec:LM_results}. Lastly, Section \ref{ssec:final_results} simulates the final gains on a new ASR application.

\subsection{Confidence modeling}\label{ssec:conf_results}

Various confidence models including a normalization of the token posteriors and an utterance-level calibration, as described in Section \ref{ssec:conf}, have been tried for data selection. For each confidence model, the confidence filter settings have been optimized as will be explained in Section \ref{ssec:AM_results}. Unfortunately, our results did not indicate any AL improvement by using more sophisticated confidence models. Only marginal (below 2\% relative WER) differences not necessarily consistent across the experiments were observed. Our explanation is two-fold: First, the ranking across the utterances in the selection pool is not substantially affected by the different models. Second, even when the ranking is altered, the informativeness of the switched utterances is probably comparable, therefore not leading to any dramatic difference in recognition performance.

\begin{figure}[!htbp]
  \centering
  \includegraphics[width=0.42\textwidth, height=0.16\textwidth]{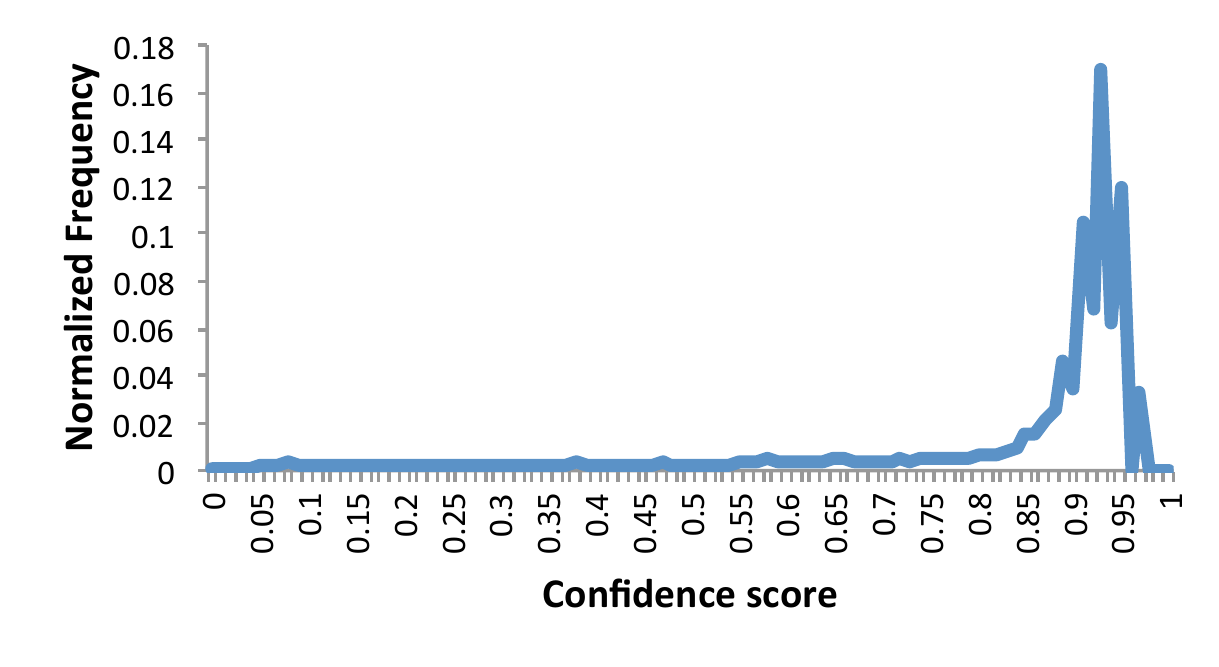}
	\vspace{-10pt}
  \caption{Histogram of the standard confidence scores.}
  \label{fig:ConfHisto}  
	\vspace{-15pt}  
\end{figure}

The rest of this paper therefore employs a simple confidence model: a polynomial is used to map the token posteriors to the observed word probabilities, which are then combined by geometrical mean. The distribution of these scores over the 550h selection pool is shown in Figure \ref{fig:ConfHisto}. Note that the various peaks in the high confidences are due to a dependency on the hypothesis length. As can be seen, the baseline model is already rather good: respectively 11.6, 19.0 and 24.0\% of the utterances have a confidence score lower than 0.5, 0.7 and 0.8.

\subsection{Impact on the AM}\label{ssec:AM_results}

In this section, we focus on the impact of AL and SST purely on the AM. The LM and the vocabulary are therefore fixed to that of the baseline. For both supervised and unsupervised data selection, our approach relies on applying a filter to the confidence scores where data is selected if the confidence score is between some given lower and upper bounds.

\subsubsection{Active learning only}\label{sssec:ALonly}

In a first stage, we optimized the filter used for supervised data selection. We varied the lower filter bound in the [0-0.1] range in order to remove possibly uninformative out-of-domain utterances. The upper bound was varied in the [0.4-0.9] range, leading to a total of 20 filters. The resulting AMs were analyzed on the development set. The main findings were that as long as the lower bound does not exceed 0.05 and the upper bound does not exceed 0.8 (which corresponds to the beginning of the main mode in Figure \ref{fig:ConfHisto}), the results were rather similar (with differences lower than 1\% relative). It seems to be important, though, not to go beyond 0.8 as this would strongly compromise the informativeness of the selected utterances. In addition, we have tried to apply utterance length filtering in cascade with the confidence-based selection. This operation however did not turn out to provide any gain.

Based on these observations, we have used the [0-0.7] confidence filter for AL data selection. When 100h of supervised data was added to the baseline, this technique reduced the WER by about 2\% relative over the RS scheme.


\subsubsection{Including unsupervised data}\label{sssec:SST}

In a second stage, we optimized the method for selecting the unsupervised data. On top of the 50h baseline set and the 50h of AL data (selected as mentioned in Section \ref{sssec:ALonly}) we added unsupervised data selected according to different confidence filters and analyzed again the AM performance after XE training on the development set.
\begin{figure}[!htbp]
  \vspace{-8pt}
  \centering
  \includegraphics[width=0.48\textwidth]{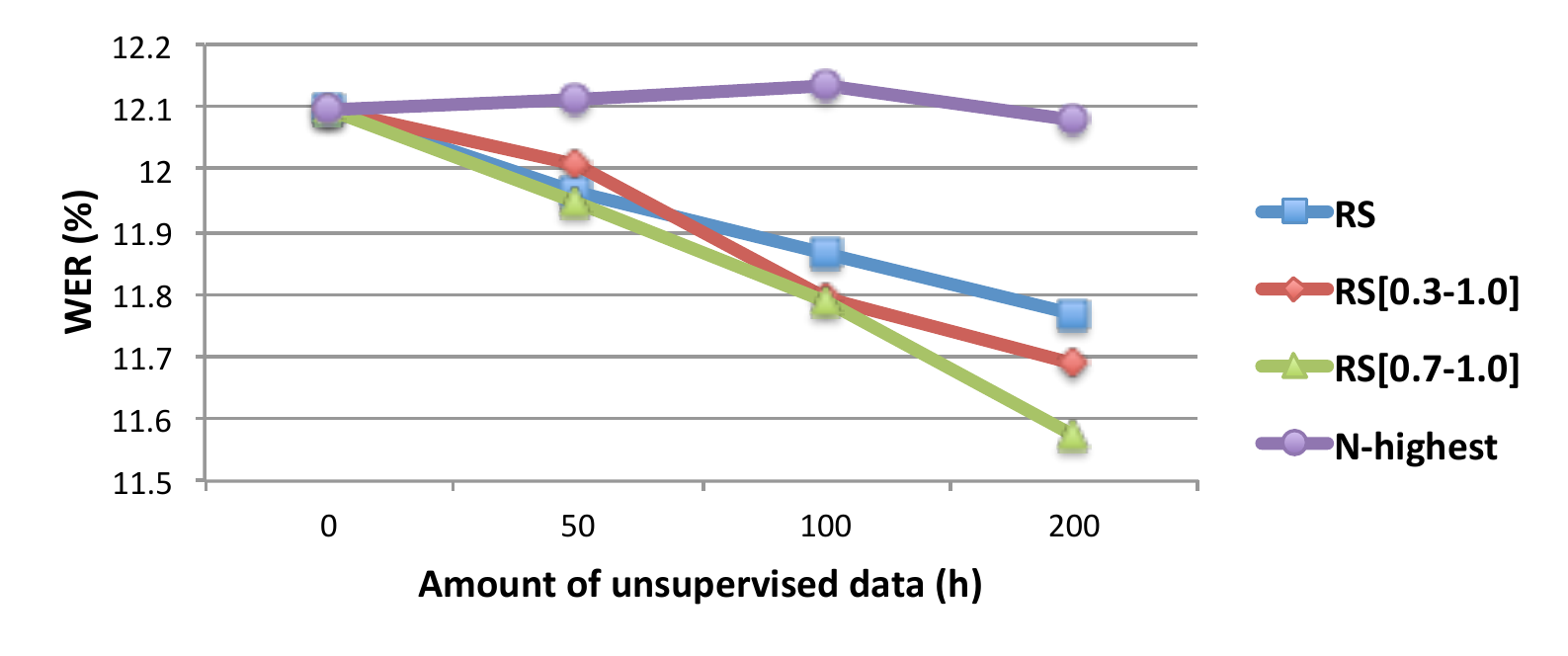}
	\vspace{-25pt}
  \caption{Benefits of unsupervised data on a XE-trained AM.}
  \label{fig:ALonly_unsup}  
	\vspace{-15pt}  
\end{figure}

Our attempts to integrate utterance length and frequency filtering as in \cite{Kapralova} were not conclusive as no significant gains were obtained. We also remarked a slight degradation if the upper bound for confidence filtering does not reach the limit of 1.0.  We therefore focused on pure confidence filtering with an upper bound of 1.0 in the remainder of our experiments. The plot in Figure \ref{fig:ALonly_unsup} compares 4 techniques of unsupervised data selection: unfiltered random sampling (\emph{RS}), confidence filtering using two different confidence filters (\emph{RS[0.3-1.0]} and \emph{RS[0.7-1.0]}), and choosing the sentences with the highest confidence scores (\emph{N-highest}). We obtained the best results with the [0.7-1.0] confidence filter. The poor performance of the N-highest scores approach can be explained by the fact that it just adds high confidence utterances which contain little new information. On the other hand, with a low lower bound of the confidence filter (as in [0.3-1.0] or \emph{RS}) the label quality becomes worse and the results also degrade. A remarkable fact is that the more unsupervised data, the better the performance of the AM. The addition of 200h of unsupervised data yielded an improvement of 4.5\% relative. The same experiment was replicated with 100h of AL data, and the conclusions remained similar, except that the gain reached 3.5\% (and not 4.5\%) this time.

\subsection{Impact on the LM}\label{ssec:LM_results}

The most important component of the interpolated LM is the one trained on transcriptions of the in-domain utterances. In this section we study the impact of different methods to select in-domain data and add it to this component on top of the 50h of the baseline model. All other LM components are kept constant. We consider three data pools from which training data could be taken: supervised data from the 100h AL data pool which was selected using the [0-0.7] confidence filter as described in Section \ref{sssec:ALonly}, supervised data from the complete pool of 550h, and unsupervised data from the same 550h pool, taken from the first hypothesis of the ASR results of the baseline model.

\subsubsection{Perplexity results}\label{sssec:Perplexity}

\begin{figure}[!b]
  \vspace{-8pt}
  \centering
  \includegraphics[width=0.48\textwidth]{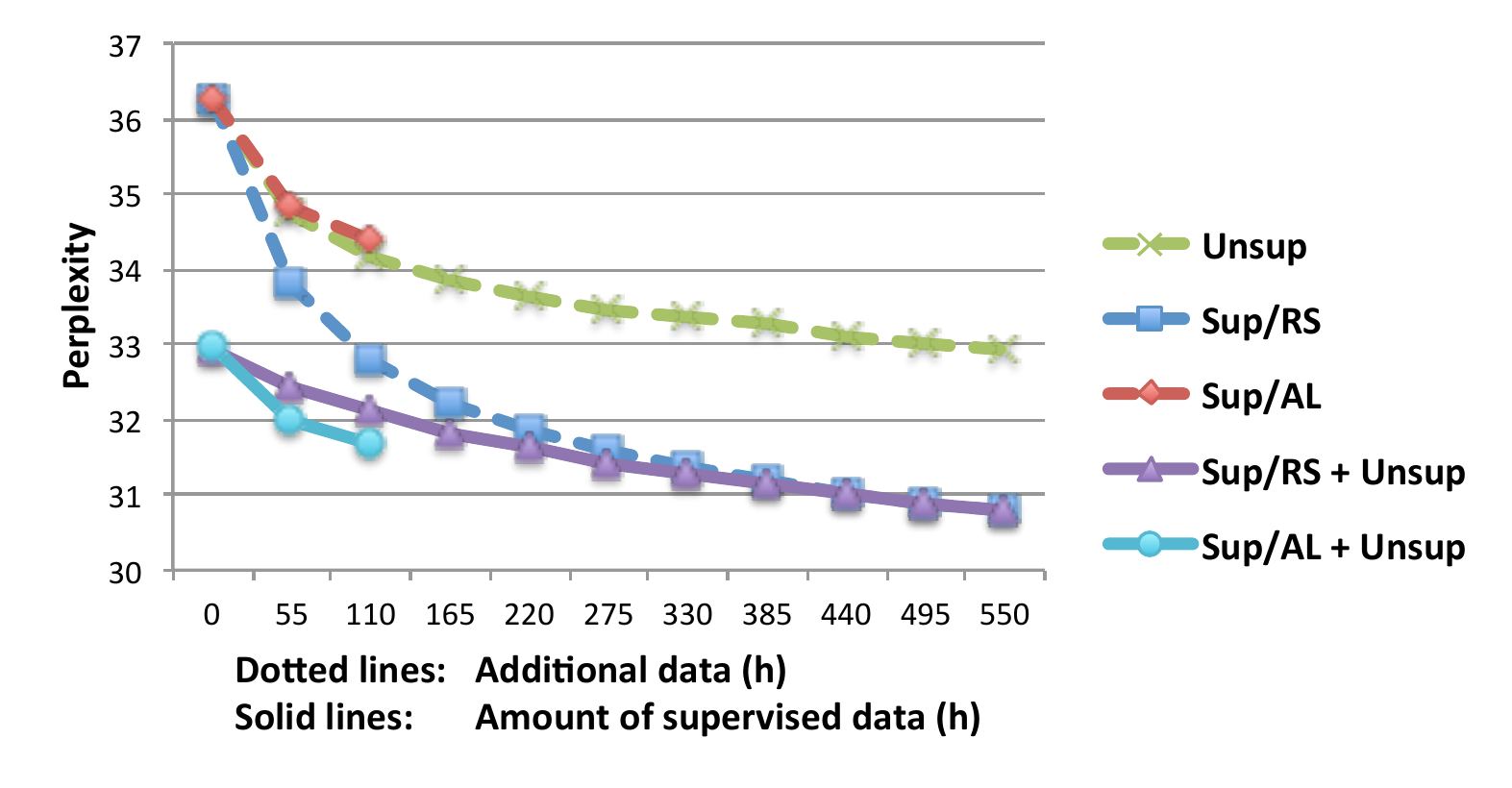}
	\vspace{-25pt}
  \caption{LM perplexity for different types of data.}
  \label{fig:Perplexity}  
\end{figure}

In a first experiment we calculated perplexities when an increasing amount of data was added to the LM. Since perplexity values are hard to compare when using different vocabularies, we kept the vocabulary fixed to that of the baseline. The dotted lines in Figure \ref{fig:Perplexity} show the perplexities if data is randomly sampled from the supervised data (\emph{Sup/RS}), if the data is sampled from the recognition results (\emph{Unsup}), and if the data is sampled from the the AL data pool (\emph{Sup/AL}). It can be seen that adding more application data improves the model irrespective of the source. Already just adding unsupervised data gives a big perplexity reduction from 36.3 to 33.0. However, there is a significant gap between the supervised (\emph{Sup/RS}) and the unsupervised (\emph{Unsup}) case. Adding just the AL data does not perform as well as random sampling from the complete pool. On one hand, in this case the label quality is higher compared to the unsupervised data but on the other hand, due to the selection process, the data is no longer representative to the application. Contrary to the AM, which is discriminatively trained, the LM is a generative model which in general is much more vulnerable for missing representativeness.

In the next experiments, shown as solid lines in Figure \ref{fig:Perplexity}, we combine supervised and unsupervised data with the goal to overcome the bias in the data and to make the best use of all the data. Supervised data was again selected either by RS (\emph{Sup/RS + Unsup}) or by AL  (\emph{Sup/AL + Unsup}) but in addition, all the remaining data of the 550h data pool were used in training as unsupervised data. This way we always use the complete data and thus maintain the representativeness. The beginning of the curves correspond to 550h unsupervised data. In the case of \emph{Sup/RS + Unsup} it drops constantly to final value of 550h supervised data. Contrary to the previous experiment, when applying AL to select the training data (\emph{Sup/AL + Unsup}), we no longer suffer from a bias of the data and the model performs even slightly better than RS.

\subsubsection{Recognition results}\label{sssec:LM_reco}

\begin{figure}[!hb]
  \vspace{-8pt}
  \centering
  \includegraphics[width=0.48\textwidth]{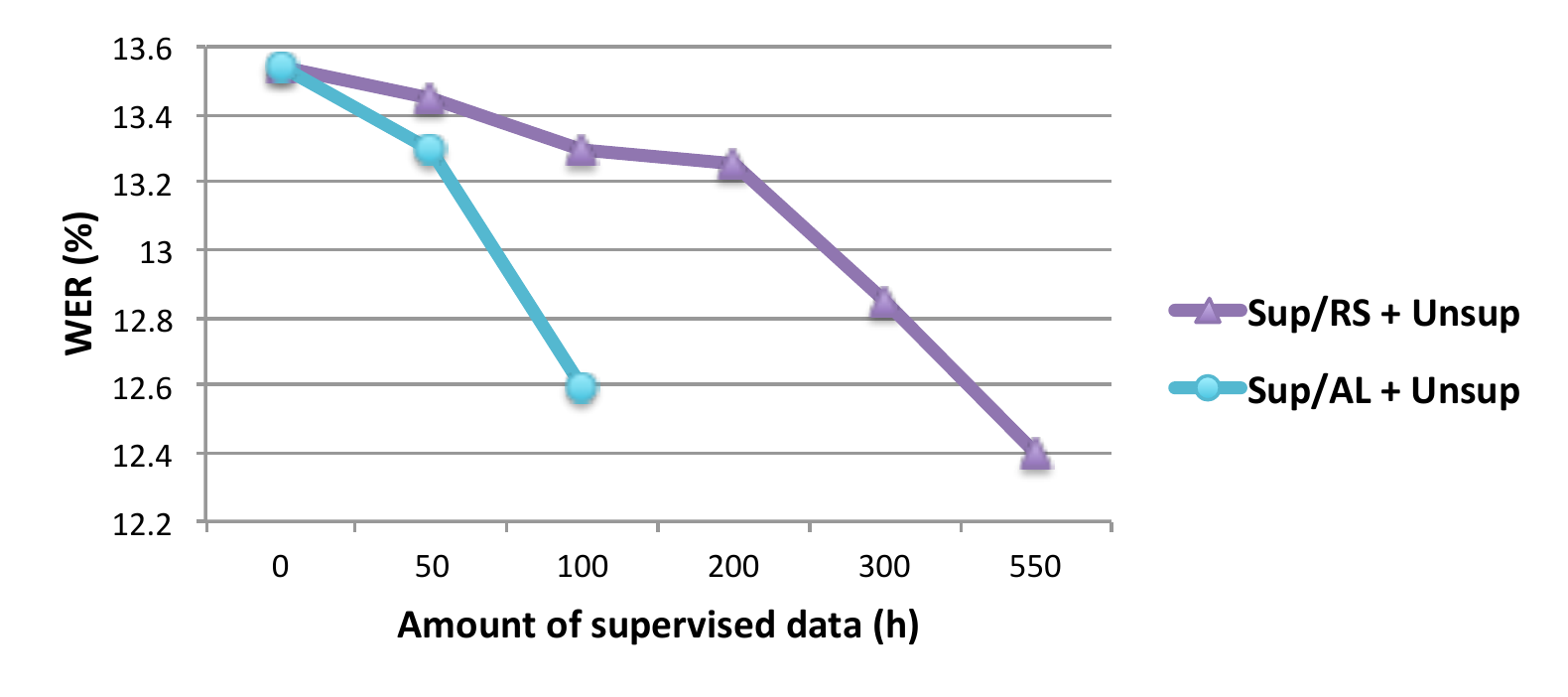}
	\vspace{-25pt}
  \caption{ASR results with updated LM and vocabulary.}
	\vspace{-15pt}  
  \label{fig:LM_reco}  
\end{figure}

It is well known that gains in perplexity do not always correspond to WER improvements. We therefore ran recognition experiments using the LMs from Section \ref{sssec:Perplexity}. Since it is beneficial to the models we always added the unsupervised data on top of the supervised data. The AM was kept fixed to the baseline. As we were no longer restricted by the perplexity measure, we also updated the vocabulary according to the selected supervised training data in these experiments. The results in Figure \ref{fig:LM_reco} show that the improvements in perplexity are also reflected in a better WER even though part of the improvements might also be due to the increased vocabulary coverage. It is interesting to observe that, when adding 100 hours of supervised data, the gains for AL are much higher than for RS. In total, the impact of AL combined with SST on LM is outstanding: after 100h of transcribed data, the gain over the RS baseline reaches 5.3\% relative. It is also worth emphasizing that 100h AL and roughly 400h RS are equivalent in terms of LM performance.

\subsection{Final results}\label{ssec:final_results}

Finally, we simulate the improvements that would be yielded in a new application by applying confidence-based AL and SST to both the AM and LM. We considered the different LMs as suggested in Section \ref{ssec:LM_results}. For AM building, we limited the unsupervised set to 200h across our experiments. For XE training, SST was applied, following the findings from Section \ref{sssec:SST}. For sequence-discriminative bMMI training, it is known that possible errors in the transcripts can have a dramatic negative influence on the quality of the resulting AM \cite{Manohar}. Therefore, two strategies were investigated: \emph{i)} considering the aggregated set of supervised and unsupervised data for bMMI training; \emph{ii)} discard any unsupervised data and only train on the supervised set. Our results indicate that the inclusion of unsupervised data led to a degradation of about 2.5\%, and this despite the relatively high lower bound used in the confidence filter (0.7). The first strategy was therefore used in the following.

\begin{figure}[!hb]
  \vspace{-8pt}
  \centering
  \includegraphics[width=0.48\textwidth]{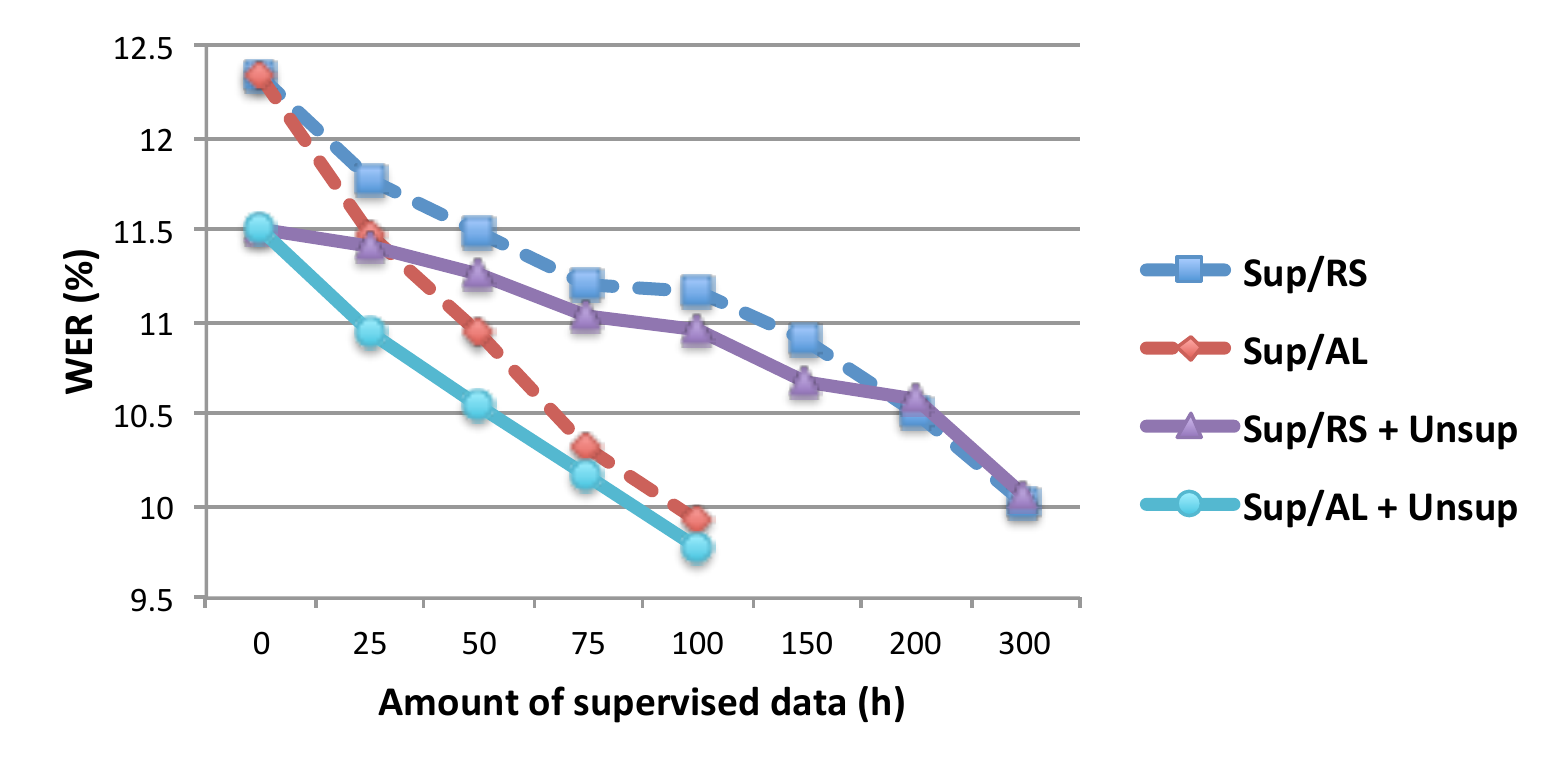}
	\vspace{-25pt}
  \caption{Final simulation: both the AM and LM are updated.}
  \label{fig:FinalPlot}  
\end{figure}

Figure \ref{fig:FinalPlot} shows the final simulation results after bMMI training. It is worth noting that the results obtained after XE training were very much in line and led to very similar improvements. Two main conclusions can be drawn from this graph. First, the unsupervised data is particularly important at the very beginning, where it allows a 6.8\% relative improvement. Nevertheless, the gains of SST vanish as more supervised data is collected. In the AL case, the advantage from SST almost completely disappears after 100h of additional supervised data. Secondly, AL carries out significant improvements over RS. It can be seen that the WER obtained with 100h of AL is comparable (even slightly better) to that using 300h of RS data, hence resulting in a reduction of the transcription budget of about 70\%. Alternatively, one can observe that, for a fixed transcription cost of 100h, AL achieves an appreciable WER reduction of about 12.5\% relative over the range of added supervised data.

\section{Conclusions}\label{sec:conclu}

This paper aimed at simulating the benefits of AL and SST in a new ASR application by applying confidence-based data selection. More sophisticated confidence models have been developed, but they did not provide any gain for training data selection for AL. Regarding AM training, AL alone was found to yield a 2\% relative improvement. Combining it with SST turned out to be essential, especially when the amount of supervised data is limited. Adding 200h of unsupervised data to 50h of AL gave a 4.5\% gain on the AM trained by cross-entropy. On the contrary, any unsupersived data was harmful to sequence-discriminative bMMI training. Beyond these improvements on the AM, combining AL and SST allowed a significant improvement (about 5\%) of the LM. Our final results indicate that applying AL to both AM and LM provides an encouraging 70\% reduction of the transcription budget over RS, and these gains seem to scale up rather well as more and more utterances are transcribed.

  \newpage
  \eightpt
  \bibliographystyle{IEEEtran}

  \bibliography{mybib}


\end{document}